\documentclass[11pt,a4paper]{article}
\usepackage[hyperref]{acl2018}
\usepackage{times}
\usepackage{latexsym}
\usepackage{lipsum}
\usepackage{graphicx}
\usepackage{float}
\usepackage{url}
\usepackage{textcomp}
\usepackage{booktabs}
\usepackage{tabularx}
\usepackage{array}
\usepackage{tikz}
\usetikzlibrary[arrows.meta,bending]
\usetikzlibrary{positioning}

\newcolumntype{L}[1]{>{\raggedright\let\newline\\\arraybackslash\hspace{0pt}}m{#1}}
\newcolumntype{C}[1]{>{\centering\let\newline\\\arraybackslash\hspace{0pt}}m{#1}}
\newcolumntype{R}[1]{>{\raggedleft\let\newline\\\arraybackslash\hspace{0pt}}m{#1}}

\aclfinalcopy 

\title{IIT (BHU) Varanasi at MSR-SRST 2018: A Language Model Based Approach for Natural Language Generation}

\author{Avi Chawla, Ayush Sharma, Shreyansh Singh and A.K. Singh\\ \\
  Indian Institute of Technology (BHU), Varanasi, India \\
  {\tt \{avi.chawla.cse16,ayush.sharma.cse16\}@iitbhu.ac.in}\\ {\tt \{shreyansh.singh.cse16,aksingh.cse\}@iitbhu.ac.in} \\}

\date{}

\begin{document}
\maketitle
\begin{abstract}
  This paper describes our submission system for the Shallow Track of Surface Realization Shared Task 2018 (SRST\textquotesingle 18). The task was to convert genuine UD structures, from which word order information had been removed and the tokens had been lemmatized, into their correct sentential form. We divide the problem statement into two parts, word reinflection and correct word order prediction. For the first sub-problem, we use a Long Short Term Memory based Encoder-Decoder approach.  For the second sub-problem, we present a Language Model (LM) based approach. We apply two different sub-approaches in the LM Based approach and the combined result of these two approaches is considered as the final output of the system.

\end{abstract}

\section{Introduction}

\begin{figure*}[t]
  \includegraphics[width=\textwidth]{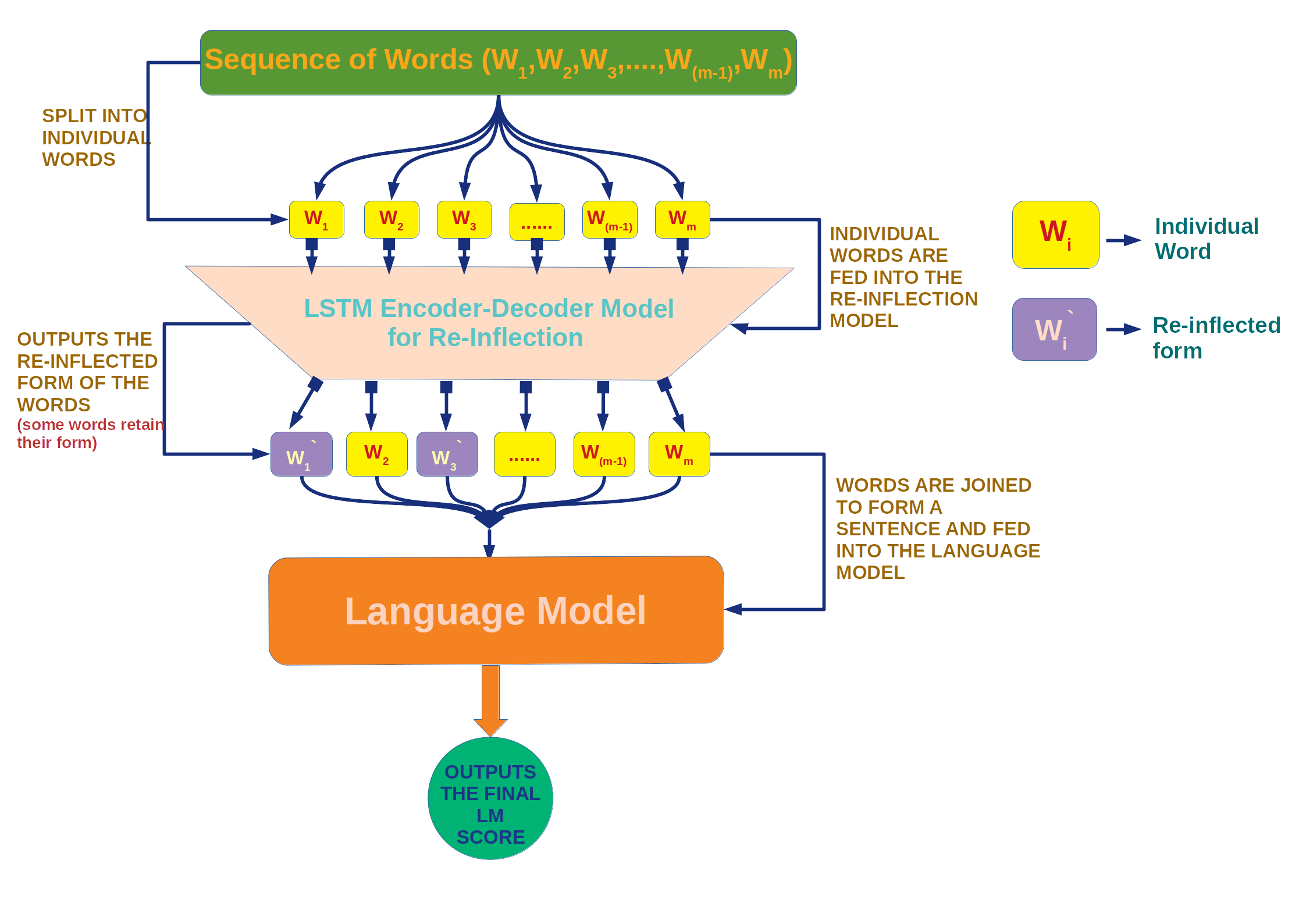}
  \caption{Architecture of the Proposed Model - Word sequence ($w_1$, $w_2$, $w_3$, ...,$w_n$) is reinfected into ($w_1^\prime$ , $w_2$, $w_3^\prime$, ..., $w_m$), where $w_i^\prime$ are the changed words due to reinfection. Final output gives the LM Score for the sequence of reinflected words. Model is run on different possible combinations and the sequence with best LM Score is chosen.}
\end{figure*}

SRST\textquotesingle 18~\cite{mille-EtAl:2018:MSR-SR}, organized under ACL 2018, Melbourne, Australia aims to re-obtain the word order information which has been removed from the UD Structures~\cite{nivre2016universal}. Universal Dependency (UD) structure is a tree representation of the dependency relations between words in a sentence of any language. Made using the UD framework, the structure of the tree is determined by the relation between a word and its dependents. Each node of this tree holds the Part of Speech (PoS) tag and morphological information as found in the original annotations of the word corresponding to that node. \\
The morphological information of a word includes the information gained from the formation of the word and its relationship with other words. Morphological information includes gender, animacy, number, mood, tense etc. \\
In this problem, we are given
\begin{enumerate}
\item Unordered dependency trees with lemmatized nodes.
\item The nodes hold PoS tags and morphological information as found in the original annotations.
\item The corresponding ordered sentences.
\end{enumerate}

Our system may find its use in various NLP applications like Natural Language Generation (NLG)~\cite{reiter1997building}. NLG is a major and relatively unexplored sub-field of NLP. Our system can be used in tasks like Question Answering, where you have the knowledge base with you which may not necessarily be holding the correct word order information but must be holding the dependencies between the words. This is where NLG is useful, where you take all the dependencies available with you and try to generate language from it which can be understood and interpreted easily by the person or user. This system also finds its application in other important tasks like abstractive text summarization~\cite{barzilay2005sentence} and image caption generation~\cite{xu2015show}, since having the correct word order is a must for any text.\\
Our system makes use of a Long Short Term Memory (LSTM) ~\cite{Hochreiter:1997:LSM:1246443.1246450} based Encoder-Decoder~\cite{K17-2007} approach to tackle the subproblem-1 of this track, i.e word re-inflection and then we make use of a probabilistic and statistical Language Model to determine the correct word order from the unordered sentences. Statistical Language Modeling, or Language Modeling or LM in short, is a technique which uses probabilistic models that are able to predict the next word in the sequence given the words that precede it. This is done by assigning a probability to the whole sequence.\\ 
The shared task organizers provided the training and a small development dataset for building our systems. A period of about 3 weeks was given for submitting our predictions on the test set.

The rest of the paper is structured as follows. Section 2 discusses, in brief, the dataset for the task. Section 3 explains our proposed approach in detail. We discuss what models we have used to re-inflect the words and generate ordered sentences from the jumbled sentences. Section 4 explains how the system is evaluated and Section 5 states the results we have obtained. We have also included an analysis of our system in Section 6. We conclude our paper and discuss its future prospects in Section 7.

\section{Data}

We used the dataset provided by the shared task organizers for training our system. No other external dataset was used in training. The dataset of the shared task is comprised of two sets of files, a .conll file containing the UD structures of sentences, and a text file containing the ordered sentences along with their sentence ids. We have worked only on the English language dataset.
There are around 12000 sentences in the training file and approximately 3000 sentences in the development file. The complete details of the dataset can be found here\footnote{\url{http://taln.upf.edu/pages/msr2018-ws/SRST.html\#data}}.

\section{Proposed System}

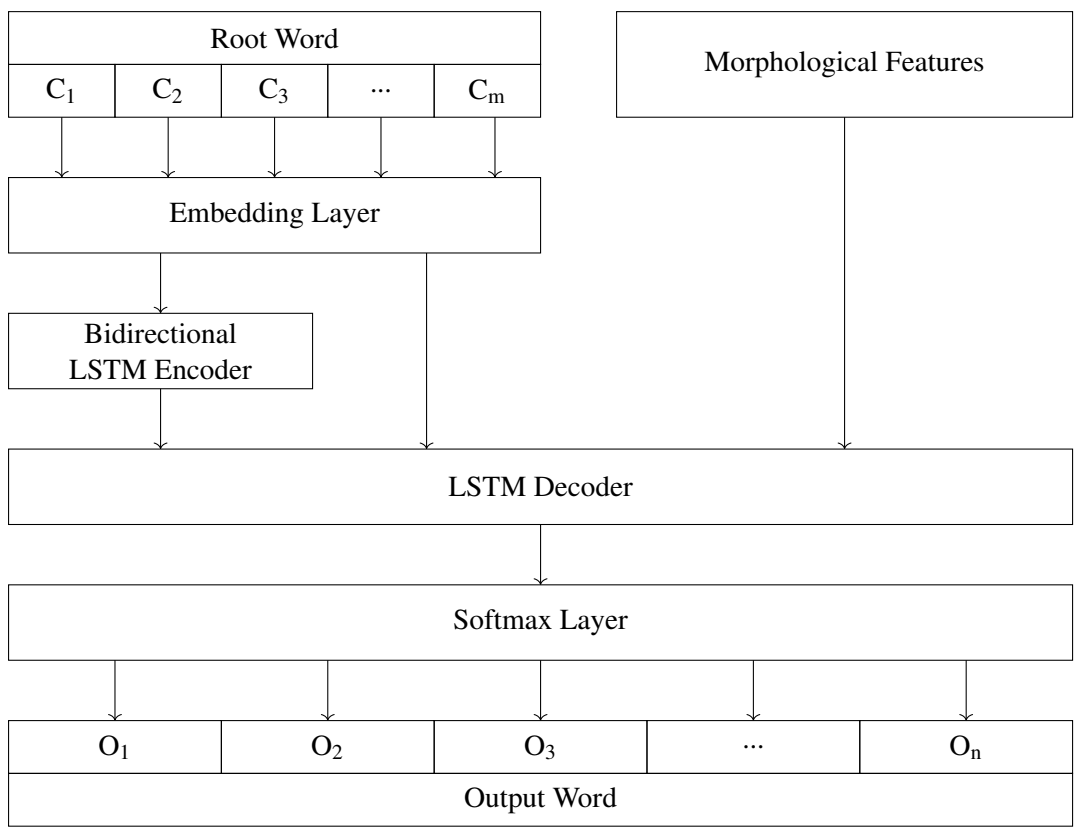
\begin{figure*}[h!]
  \begin{tikzpicture}

    \draw (0,0) rectangle (0,0);
    \draw (1,0) rectangle (15,0.7) node[midway] {Output Word};
    \draw (1,0.7) rectangle (3.8,1.4) node[midway] {O\textsubscript{1}};
    \draw (3.8,0.7) rectangle (6.6,1.4) node[midway]  {O\textsubscript{2}};
    \draw (6.6,0.7) rectangle (9.4,1.4) node[midway]  {O\textsubscript{3}};
    \draw (9.4,0.7) rectangle (12.2,1.4) node[midway]  {...};
    \draw (12.2,0.7) rectangle (15,1.4) node[midway]  {O\textsubscript{n}};

    \draw (1,2.2) rectangle (15,3.2) node[midway]  {Softmax Layer};

    \coordinate (A) at (2.4, 1.4);
    \coordinate (B) at (5.2, 1.4);
    \coordinate (C) at (8.0, 1.4);
    \coordinate (D) at (10.8, 1.4);
    \coordinate (E) at (13.6, 1.4);
    \coordinate (F) at (2.4, 2.2);
    \coordinate (G) at (5.2, 2.2);
    \coordinate (H) at (8.0, 2.2);
    \coordinate (I) at (10.8, 2.2);
    \coordinate (J) at (13.6, 2.2);
    \draw[->] (F) -- (A);
    \draw[->] (G) -- (B);
    \draw[->] (H) -- (C);
    \draw[->] (I) -- (D);
    \draw[->] (J) -- (E);

    \draw (1, 4.0) rectangle (15, 5.0) node[midway] {LSTM Decoder};

    \coordinate (K) at (8, 3.2);
    \coordinate (L) at (8, 4.0);
    \draw[->] (L) -- (K);

    \draw (1, 5.8) rectangle (5, 6.8) node[midway, align = center] {Bidirectional \\ LSTM Encoder};

    \coordinate (M) at (3, 5.8);
    \coordinate (N) at (3, 5.0);
    \draw[->] (M) -- (N);

    \draw (1,7.6) rectangle (8, 8.6) node[midway] {Embedding Layer};

    \coordinate (A3) at (3, 7.6);
    \coordinate (A4) at (3, 6.8);
    \draw[->] (A3) -- (A4);

    \coordinate (O) at (6.5, 7.6);
    \coordinate (P) at (6.5, 5.0);
    \draw[->] (O) -- (P);

    \draw (1,9.4) rectangle (2.4,10.1) node[midway] {C\textsubscript{1}};
    \draw (2.4,9.4) rectangle (3.8,10.1) node[midway] {C\textsubscript{2}};
    \draw (3.8,9.4) rectangle (5.2,10.1) node[midway] {C\textsubscript{3}};
    \draw (5.2,9.4) rectangle (6.6,10.1) node[midway] {...};
    \draw (6.6,9.4) rectangle (8.0,10.1) node[midway] {C\textsubscript{m}};
    \draw (1, 10.1) rectangle (8.0, 10.8) node[midway] {Root Word};

    \coordinate (Q) at (1.7, 9.4);
    \coordinate (R) at (3.1, 9.4);
    \coordinate (S) at (4.5, 9.4);
    \coordinate (T) at (5.9, 9.4);
    \coordinate (U) at (7.4, 9.4);
    \coordinate (V) at (1.7, 8.6);
    \coordinate (W) at (3.1, 8.6);
    \coordinate (X) at (4.5, 8.6);
    \coordinate (Y) at (5.9, 8.6);
    \coordinate (Z) at (7.4, 8.6);
    \draw[->] (Q) -- (V);
    \draw[->] (R) -- (W);
    \draw[->] (S) -- (X);
    \draw[->] (T) -- (Y);
    \draw[->] (U) -- (Z);

    \draw (9, 9.4) rectangle (15, 10.8) node[midway] {Morphological Features};

    \coordinate (A1) at (12, 9.4);
    \coordinate (A2) at (12, 5.0);
    \draw[->] (A1) -- (A2);

   \end{tikzpicture}
  
  \caption{Architecture of the Word Re-inflection model - \textit{$C_1$, .., $C_m$} represent characters of the root word while \textit{$O_1$, ..,$O_n$} represent characters of the output word.}

\end{figure*}

In the Shallow Track of the shared task, we had two subproblems to deal with. First one was the re-inflection problem and the second one involved the task of re-obtaining the correct word order from the unordered UD structure.
We shall refer to these problems as Subproblem-1 and Subproblem-2 subsequently in this paper. Subproblem-1 is the word re-inflection problem and Subproblem-2 is the word ordering problem.\\
The complete architecture of the proposed model is shown in Figure 1.


\subsection{Sub Problem-1: Word Re-inflection}

In the given UD structure, the words are given in lemmatized form. Before proceeding to determine the correct order of words, these lemmatized words must be re-inflected to convert them into their correct form. For the task of re-inflection, we implemented an LSTM based encoder-decoder model. The morphological information is given in CoNLL format. Since majority of the past work in reinfection uses the UniMorph annotation format of the morphological features, we first converted our morphological features from CoNLL to an approximation of the UniMorph format by modeling a manual mapping between the two tagsets. Eg. For the word ``preacher'', the CoNLL annotation format is Noun \& Number=Sing. We convert this to N;SING. This can be treated as an approximation of the UniMorph annotation format, which is sufficient for us.\\ \\ This approach is based on a submission in the CoNLL-SIGMORPHON-2017 Shared Task~\cite{K17-2007}. The model takes into account the fact that the root word (lemmatized form) and the target word (re-inflected form) are similar except for the parts that have been changed due to re-inflection. The model outputs the target word character by character, thus handling both the cases when there are prefix or suffix changes (\textit{play} to \textit{playing}) or changes occurring in the middle of the word (\textit{man} to \textit{men}).\\
The root word is represented using character indices, while the associated morphological features are represented in the form of a binary vector. A root word embedding for each word is formed by making a 64 dimensional character embedding of each character. This embedding is fed into a bidirectional LSTM encoder. The output of this encoder, along with the root word embedding and the binary vector representing the morphological features, acts as input to the LSTM decoder. A softmax layer is then used to predict the character at each position of the output word. To maintain a common length for all words, a padding of 0 is used. The architecture of this model is shown in Figure 2.

\subsection{Sub Problem-2: Word-Ordering}
We have used a probabilistic and statistical Language Model to tackle this subproblem. After re-inflecting the words in the UD-Structures, the next step is to obtain the correct word-order of each sentence. For this, we make use of the SRILM Toolkit~\cite{Stolcke02srilm--}. 

Before predicting the correct word-order, we follow the following steps to train the Language Model:
\begin{enumerate}
\item We generate a vocabulary file from the corpus of ordered sentences. The vocab file is the list of all unique words occurring in the corpus, with each word in a different line.
\item After we have the vocab file with us, we make use of this and the ordered sentence data to generate a .lm file using the SRILM toolkit. This file contains the probability scores of the associated n-grams (till trigrams) present in the corpus.
\end{enumerate}

After calculating these probabilities, we move on to solve the prime objective of this subproblem, which is to find the correct word order of the unordered sentences.

For this, we have used two methods. Selecting which method to use depends on the sentence length. 
\begin{itemize}
\item Method 1: 4-gram LM Based Approach
\item Method 2: Variable n-gram LM Based Approach
\end{itemize}

Method 1 is used in cases where the sentence length is more than 23 (23 being a hyperparameter in this case) and Method 2 is used for sentences having their length less than or equal to 23. Note that we have predicted the sentences without any punctuations in it. All the punctuations appearing in a sentence were removed.  However, a full stop was added at the end of each predicted sentence.

We thoroughly describe the two methods below.

\subsubsection{Method 1: 4-gram LM Based Approach}
This method is used to find the correct sentential form of those sentences which have length greater than 23. We define the Language Model score (LM score) of a string to be the probability measure of that string being drawn from some vocabulary. If the vocabulary is made using linguistically correct sentences, then a higher Language Model score indicates higher probability of a sentence being linguistically correct. An ideal approach would be to calculate the LM score of all possible permutations of all the words in  a sentence and select the permutation with the highest LM score. Since this is computationally very expensive (specially for large sentences), hence we check for permutations of at most 4 words only. If the sentence length is less than or equal to 4, we select the permutation with the highest LM score. For length greater than 3, we calculate the LM score of all the possible 4-grams for the given sentence. Then, we select the one which gives the highest LM score and choose this as the start of the sentence sequence. For determining rest of the sequence, we follow the following steps:
\begin{enumerate}
\item Maintain a list of remaining words (LRW). This list consists of all the words in the sentence, except the 4 words which have been selected as the start of the sentence sequence.
\item Repeat the following until no word is left in LRW:
	\begin{itemize}
    \item For each word left in LRW, check which word, on addition to the predicted sequence gives the maximum Language Model Score. Let this word be \textit{w}.
	\item Add \textit{w} to the predicted sequence and remove it from LRW.
	\end{itemize}
\end{enumerate}

By following the above mentioned steps, we get the final sequence of words as predicted by Method-1 of our LM approach.

\subsubsection{Variable N-gram LM Based Approach}

This method was used to find the correct sentential form of those sentences having length less than or equal to 23. In this method, instead of only looking for the best 4-gram, we look for various bigrams and trigrams as well. For example, for a sentence of length 6, we break the sentence into (3-gram, 2-gram, 1-gram), (2-gram, 2-gram, 2-gram) and (3-gram, 3-gram). Similarly, we have manually broken each sentence length into different combination of unigrams, bigrams and trigrams. We calculate the LM score of different relative arrangements of these n-grams. Out of all the possible relative arrangements, the one which gives the maximum LM Score is chosen as the prediction of our model for that jumbled sentence.\\
The idea behind choosing different combinations of n-grams is that a sentence is generally divided into different chunks and if we are able to identify the chunks in which the words of a sentence appear, we can then use a language model to find which possible sequence would have been the best out of all the different possible relative arrangements of these chunks of words.

\section{Evaluation}

\textbf{Cross Validation (CV): }We trained our model on the training data and did predictions on the development data, both of which were provided by the shared task organizers. These predictions were considered as the CV Score of our model. The metrics that were used to evaluate the model were BLEU~\cite{Papineni:2002:BMA:1073083.1073135}, NE DIST and NIST~\cite{Doddington:2002:AEM:1289189.1289273}. Evaluation script for the same was also provided by the organizers.

\textbf{Test: }Once we were done with the optimal tuning of our model using the CV score, we used our model to generate ordered sentences on the test data. We trained on the full training data for the re-inflection task and combined the training and development data to generate the language model (.lm) file for the word-ordering task.

\section{Results}

We worked on Track 1 (Shallow track) of the shared task for the English language. The performances of our system, the system which scored the highest for English and the system which scored the highest when averaged over the scores of all the languages is given in the table below. Evaluation is done across various metrics provided by the shared task organizers. Note that all the scores given below are for English language only.

\begin{table}[h!]
\fontsize{8.5}{10.5}\selectfont
\begin{center}
\begin{tabular}{|l|c|c|c|} 
 \hline
\textbf{} & \textbf{BLEU Score} & \textbf{NE DIST} & \textbf{NIST}\\
 \hline
IIT (BHU) Varanasi & 8.04 & 47.63 & 7.71\\
Highest for English & 69.14 & 80.42 & 12.02\\
Highest Average & 55.29 & 65.9 & 9.58\\ 
 \hline
\end{tabular}
\end{center}
\caption{Scores for English on test data.}
\end{table}


For word reinflection, the LSTM based encoder-decoder model correctly predicted the reinflected forms of 95.8\% words when trained on the training dataset and tested on the development dataset.

\section{Analysis}

Our model for the word reinflection sub-problem produces good results. But, the results for the word reordering sub-problem are not very good. Total 8 teams submitted their systems in the shared task, and our system was ranked the last for English by each of the three metrics given in the Results section. Some of the reasons for this are

\begin{itemize}
  \item The sentences submitted as output did not have punctuations except a full stop at the end. Because of this, our sentences lacked punctuations occurring inside a sentence. Also, it is not necessary that a sentence ends with a full stop only.
  \item Enumerating over permutations for the word reordering sub-problem was computationally expensive for sentences of length greater than 23. So, we had to use the 4-gram approach for such sentences, which does not perform that well as the variable n-gram approach. Since there were many sentences having length greater than 23 in the test set, the overall results got adversely affected. For example, ``\textit{It looks pretty cool on the other hand.}'' is a prediction by our model, which is a decent sentence. However, the prediction ``\textit{There have been the us soldiers with have to either even long since by arab fundamentalists local sunni radicals become remain or or relations sunnis committed nationalism roiled falluja tense.}'', which is 30 words long, does not appear to be a meaningful English sentence.
  \item There is another important point worth noticing with respect to the evaluation metrics. For a given set of words, there may be more than one linguistically correct word order. For example, both the sentences ``\textit{The boy reads a book.}'' and ``\textit{The book a boy reads.}'' are made up of the same set of words and both are linguistically correct as well. So, the algorithms used for evaluation of results may not give the highest possible score to a sentence even if it is linguistically correct.

\end{itemize}

\section{Conclusion and Future Work}

In this paper, we described a system which treats reinflection and word reordering as two independent sub-problems. We have used an LSTM based approach to solve the problem of re-inflection. The LSTM model works on character embeddings and predicts the re-inflected word character by character. We have observed that this type of model can be more effective and beneficial than other elementary approaches like String Matching~\cite{DBLP:journals/corr/CotterellKSWVXF17} etc.

For the Word-Ordering problem, we have worked with only statistical and probabilistic approaches till now and haven\textquotesingle t yet incorporated any deep learning based approach in our model.
Neural models are state of the art in nearly all Natural Language Processing tasks and have always performed better than statistical and probabilistic approaches. So in future, we wish to experiment with deep learning based approaches as well. One major information we didn't exploit is the dependency relations between the words. We hope to come up with a method to somehow quantify those relations and use those values to calculate an improvised language model score. Also, since a dependency tree can be interpreted as a graph, using graph matching and searching techniques is another dimension we can explore.

\bibliography{acl2018}

\begin{thebibliography}{11}
\expandafter\ifx\csname natexlab\endcsname\relax\def\natexlab#1{#1}\fi

\bibitem[{Barzilay and McKeown(2005)}]{barzilay2005sentence}
Regina Barzilay and Kathleen~R McKeown. 2005.
\newblock Sentence fusion for multidocument news summarization.
\newblock \emph{Computational Linguistics}, 31(3):297--328.

\bibitem[{Cotterell et~al.(2017)Cotterell, Kirov, Sylak{-}Glassman, Walther,
  Vylomova, Xia, Faruqui, K{\"{u}}bler, Yarowsky, Eisner, and
  Hulden}]{DBLP:journals/corr/CotterellKSWVXF17}
Ryan Cotterell, Christo Kirov, John Sylak{-}Glassman, G{\'{e}}raldine Walther,
  Ekaterina Vylomova, Patrick Xia, Manaal Faruqui, Sandra K{\"{u}}bler, David
  Yarowsky, Jason Eisner, and Mans Hulden. 2017.
\newblock \href {http://arxiv.org/abs/1706.09031} {Conll-sigmorphon 2017 shared
  task: Universal morphological reinflection in 52 languages}.
\newblock \emph{CoRR}, abs/1706.09031.

\bibitem[{Doddington(2002)}]{Doddington:2002:AEM:1289189.1289273}
George Doddington. 2002.
\newblock \href {http://dl.acm.org/citation.cfm?id=1289189.1289273} {Automatic
  evaluation of machine translation quality using n-gram co-occurrence
  statistics}.
\newblock In \emph{Proceedings of the Second International Conference on Human
  Language Technology Research}, HLT '02, pages 138--145, San Francisco, CA,
  USA. Morgan Kaufmann Publishers Inc.

\bibitem[{Hochreiter and
  Schmidhuber(1997)}]{Hochreiter:1997:LSM:1246443.1246450}
Sepp Hochreiter and J\"{u}rgen Schmidhuber. 1997.
\newblock \href {https://doi.org/10.1162/neco.1997.9.8.1735} {Long short-term
  memory}.
\newblock \emph{Neural Comput.}, 9(8):1735--1780.

\bibitem[{Mille et~al.(2018)Mille, Belz, Bohnet, Graham, Pitler, and
  Wanner}]{mille-EtAl:2018:MSR-SR}
Simon Mille, Anja Belz, Bernd Bohnet, Yvette Graham, Emily Pitler, and Leo
  Wanner. 2018.
\newblock The {F}irst {M}ultilingual {S}urface {R}ealisation {S}hared {T}ask
  ({SR}'18): {O}verview and {E}valuation {R}esults.
\newblock In \emph{Proceedings of the 1st Workshop on Multilingual Surface
  Realisation (MSR), 56th Annual Meeting of the Association for Computational
  Linguistics ({ACL})}, pages 1--10, Melbourne, Australia.

\bibitem[{Nivre et~al.(2016)Nivre, de~Marneffe, Ginter, Goldberg, Hajic,
  Manning, McDonald, Petrov, Pyysalo, Silveira et~al.}]{nivre2016universal}
Joakim Nivre, Marie-Catherine de~Marneffe, Filip Ginter, Yoav Goldberg, Jan
  Hajic, Christopher~D Manning, Ryan~T McDonald, Slav Petrov, Sampo Pyysalo,
  Natalia Silveira, et~al. 2016.
\newblock Universal dependencies v1: A multilingual treebank collection.
\newblock In \emph{LREC}.

\bibitem[{Papineni et~al.(2002)Papineni, Roukos, Ward, and
  Zhu}]{Papineni:2002:BMA:1073083.1073135}
Kishore Papineni, Salim Roukos, Todd Ward, and Wei-Jing Zhu. 2002.
\newblock \href {https://doi.org/10.3115/1073083.1073135} {Bleu: A method for
  automatic evaluation of machine translation}.
\newblock In \emph{Proceedings of the 40th Annual Meeting on Association for
  Computational Linguistics}, ACL '02, pages 311--318, Stroudsburg, PA, USA.
  Association for Computational Linguistics.

\bibitem[{Reiter and Dale(1997)}]{reiter1997building}
Ehud Reiter and Robert Dale. 1997.
\newblock Building applied natural language generation systems.
\newblock \emph{Natural Language Engineering}, 3(1):57--87.

\bibitem[{Stolcke(2002)}]{Stolcke02srilm--}
Andreas Stolcke. 2002.
\newblock Srilm -- an extensible language modeling toolkit.
\newblock In \emph{IN PROCEEDINGS OF THE 7TH INTERNATIONAL CONFERENCE ON SPOKEN
  LANGUAGE PROCESSING (ICSLP 2002}, pages 901--904.

\bibitem[{Sudhakar and Singh(2017)}]{K17-2007}
Akhilesh Sudhakar and Anil~Kumar Singh. 2017.
\newblock \href {https://doi.org/10.18653/v1/K17-2007} {Experiments on
  morphological reinflection: Conll-2017 shared task}.
\newblock In \emph{Proceedings of the CoNLL SIGMORPHON 2017 Shared Task:
  Universal Morphological Reinflection}, pages 71--78. Association for
  Computational Linguistics.

\bibitem[{Xu et~al.(2015)Xu, Ba, Kiros, Cho, Courville, Salakhudinov, Zemel,
  and Bengio}]{xu2015show}
Kelvin Xu, Jimmy Ba, Ryan Kiros, Kyunghyun Cho, Aaron Courville, Ruslan
  Salakhudinov, Rich Zemel, and Yoshua Bengio. 2015.
\newblock Show, attend and tell: Neural image caption generation with visual
  attention.
\newblock In \emph{International Conference on Machine Learning}, pages
  2048--2057.

\end{thebibliography}
\bibliographystyle{acl_natbib}

\end{document}